# Physics-Informed Deep Learning Model for Cross-Modality Super-Resolution in Fluorescence Microscopy

**MOHAMMAD SOLTANINEZHAD[1,2], ELENA CORBETTA[1,2], FRANCISCO PAEZ LARIOS[3,4], PAUL M. JORDAN[5,6], OLIVER WERZ[5,6], CHRISTIAN EGGELING[3,4,6] AND THOMAS BOCKLITZ[1,2]***

*1. Department "Photonic Data Science", Leibniz Institute of Photonic Technology, Member of Leibniz Health Technologies, Member of the Leibniz Center for Photonics in Infection Research (LPI), Jena, Germany*

*2. Work group "Photonic Data Science", Institute of Physical Chemistry (IPC) and Abbe Center of Photonics (ACP), Friedrich Schiller University Jena, Member of the Leibniz Center for Photonics in Infection Research (LPI), Jena, Germany*

*3. Institute of Applied Optics and Biophysics, Friedrich Schiller University Jena, Jena, Germany*

*4. Leibniz Institute of Photonic Technologies Department of Biophysical Imaging, Jena, Germany*

*5. Department of Pharmaceutical/Medicinal Chemistry, Institute of Pharmacy, Friedrich Schiller University Jena, 07743 Jena, Germany*

*6. Jena Center for Soft Matter (JCSM), Friedrich Schiller University Jena, 07743 Jena, Germany*

*Corresponding author: Thomas Bocklitz
thomas.bocklitz@uni-jena.de

## Abstract:

Cross-modality image translation offers a route to super-resolution fluorescence microscopy from low-resolution images while reducing phototoxicity and instrumentation demands. However, purely data-driven models can produce visually plausible outputs that are inconsistent with optical image formation. Here, we propose a physics-informed generative adversarial network for confocal-to-STED image translation that incorporates microscope-specific point spread function information into the training objective. Simulated and experimentally measured PSFs were evaluated using a limited paired confocal–STED dataset of TOM20-labeled mitochondria in human primary M2 macrophages acquired across different experimental days. Performance was assessed using reference-based and non-reference-based image-quality metrics, together with complementary frequency- and distribution-sensitive analyses. The no-reference metrics probed physics-relevant image properties, including spatial-frequency content, contrast, and signal-to-noise behavior. PSF-guided models improved structural fidelity, reduced local deviations, and achieved closer agreement with STED references than non-PSF baselines, particularly in frequency-domain analyses. These results demonstrate that optical priors can improve the structural fidelity and physical plausibility of generative microscopy models for cross-modality super-resolution imaging.

## Introduction

Fluorescence microscopy is one of the most important imaging tools in modern cell biology because it enables molecularly specific visualization of subcellular organization in intact specimens. However, conventional far-field imaging remains constrained by diffraction, which limits the direct observation of nanoscale structures (1-5). The emergence of super-resolution methods, particularly stimulated emission depletion (STED) microscopy, fundamentally changed this landscape by pushing fluorescence imaging beyond the diffraction barrier and enabling direct access to structural details that remain blurred in confocal microscopy (1–4). Despite these advantages, STED imaging typically requires specialized instrumentation, careful alignment, higher light doses, and stricter acquisition conditions, which can increase experimental complexity and aggravate photobleaching and phototoxicity (3–5). For many biological applications, this creates a persistent tension between the desire for higher spatial resolution and the practical need for gentle, routine, and scalable imaging (4,5). As a result, there is strong motivation to develop computational methods to infer super-resolved image content from more accessible acquisitions such as confocal microscopy (6–8).

The rise of deep learning has made this objective increasingly realistic (9,10). In computer vision, generative modeling and image-to-image translation frameworks established a powerful paradigm for learning mappings between related image domains. Conditional adversarial learning and cycle-consistent adversarial learning quickly influenced biomedical imaging by providing flexible solutions for paired and unpaired translation problems (9–12). In parallel, microscopy-specific studies showed that neural networks can enhance resolution, denoise images, restore optical sectioning, and reconstruct microscopy data without hardware changes, establishing deep learning as a central component of computational microscopy (13–15). At the same time, work on inverse problems and computational imaging clarified that reconstruction quality depends strongly on how much of the underlying forward model is retained in the learning pipeline (15–17). This is especially relevant in microscopy, where well-understood optical principles govern image formation and cannot be treated as an arbitrary black-box mapping without risking degraded generalization or physically implausible reconstructions (15–18).

Within microscopy, cross-modality transformation has rapidly emerged as a major application of deep learning (19–21). Early studies showed that fluorescence-like information could be predicted from transmitted-light or unlabeled inputs, introducing the concepts of in silico labeling and label-free fluorescence prediction (20,21). Related work expanded these ideas toward virtual staining, computational tissue staining, three-dimensional virtual refocusing, and super-resolution reconstruction (19,24,25,29). A major milestone for fluorescence microscopy was the demonstration that deep learning could perform cross-modality super-resolution, including confocal-to-STED transformation, suggesting that high-resolution image representations may be inferred from lower-resolution acquisitions under suitable training conditions (22). Subsequent studies extended these concepts to structured illumination microscopy, volumetric restoration, point-scanning super-resolution, and diverse biological imaging tasks (19,24,26–28). Recent reviews have further consolidated this field by showing that cross-modality transformation in microscopy now spans contrast transfer, label-free-to-fluorescence prediction, virtual staining, and resolution enhancement across paired and unpaired learning settings (23,34). Together, these studies establish cross-modality transformation as a promising strategy for reducing acquisition burden while increasing biological interpretability (19–29,34).

Nevertheless, an important limitation remains: many cross-modality models are primarily data-driven and optimize image similarity without explicitly encoding the optical image formation process (15,23,30,34). This can be problematic because visually plausible outputs are not necessarily physically faithful, and deep learning-based reconstruction models can be vulnerable to hallucinations, instability, and domain shift (18,23). In fluorescence microscopy, the point spread function (PSF) is a central descriptor of how the optical system transfers spatial information and blurs the underlying structure (6,7,30). Incorporating PSF knowledge into the learning process, therefore, provides a principled route to constrain reconstruction, preserve physically meaningful detail, and improve robustness (15,17,30,35). This broader movement toward physics-informed learning is increasingly visible across computational imaging and microscopy, from image-formation-aware deconvolution networks to recent physics-informed diffusion approaches for microscopy reconstruction (15–17,30,31,35,36). These studies support the view that combining learned representations with optical priors can yield models that are both more accurate and more trustworthy than purely empirical image translators (18,23,30–36).

In this work, we build on this direction and propose a physics-informed generative adversarial network for cross-modality super-resolution in fluorescence microscopy, focusing on the transformation from confocal to STED imaging (22,29,34,37). Using Translocase of the Outer Mitochondrial Membrane 20 (TOM20) data, we incorporate both simulated and experimentally measured PSFs into the training framework to guide the generation of physically consistent STED-like images (6,7,30,37). Importantly, the method delivers robust results even when trained on a limited dataset. This is particularly relevant for fluorescence microscopy, where acquiring large, high-quality paired datasets is often challenging. In contrast to purely appearance-driven translation, our approach is designed to preserve the structural and frequency-domain characteristics relevant to the underlying optical system (15,17,30,35). This positioning aligns with recent microscopy literature that emphasizes both the value of cross-modality transformations and the need for stronger physical constraints in learned reconstruction models (23,29,34–37). By uniting generative adversarial learning with PSF-based guidance, our study aims to advance confocal-to-STED transformation toward a more physically grounded and experimentally reliable framework for fluorescence microscopy (29,34–37).

## Results:

To assess whether PSF guidance improves translation between confocal and STED microscopy images, we modified a generative adversarial network (GAN) by incorporating the microscope point spread function (PSF) into the loss function as a physics-guided constraint. We evaluated the proposed framework on a limited paired dataset of 42 confocal and STED image pairs acquired across different experimental days, using five-fold cross-validation. In each fold, approximately 33 image pairs were used for training and the remaining pairs for testing, representing a relatively small dataset for deep learning while partially capturing day-to-day experimental variability. The analysis combined qualitative comparison of generated images with quantitative evaluation based on full-reference, no-reference, frequency-domain, and distribution-based metrics. Both simulated and experimentally measured PSFs were investigated to determine how physics-guided constraints affect translation quality. An overview of the framework and evaluation strategy is provided in Figs. 1 and 2.

### Confocal-to-STED Translation Results, Deviation Maps, and Anomaly Analysis

The PSF-guided model generated STED-like images with improved structural fidelity and local consistency compared with the Pix2Pix and CycleGAN baselines. This trend is evident in the representative confocal-to-STED translation results shown in Fig. 3, where the input confocal image, the experimentally acquired STED reference, the Pix2Pix-generated STED image, the CycleGAN-generated STED image, and the PSF-guided generated STED image are compared for two example fields of view. All AI-based models reproduced the overall mitochondrial organization observed in the STED reference while increasing the level of visible structural detail relative to the confocal input. A direct comparison shows that the PSF-guided model produced sharper and more continuous mitochondrial structures and yielded image patterns that more closely resembled experimentally acquired STED images. These differences are particularly visible in the enlarged regions, where the PSF-guided output more accurately followed the local organization and intensity distribution of the STED reference, whereas the Pix2Pix and CycleGAN outputs showed less consistent recovery of fine features.

To further assess local discrepancies, deviation maps and anomaly maps were computed with respect to the STED reference. The deviation maps indicate reduced local error for the PSF-guided model compared with the Pix2Pix and CycleGAN baselines. For sample 1, the anomaly score decreased from 0.0279 for confocal versus STED, 0.0136 for Pix2Pix versus STED, and 0.0143 for CycleGAN versus STED to 0.0131 for PSF-guided AI versus STED. For sample 2, the anomaly score decreased from 0.0740 for confocal versus STED, 0.0301 for Pix2Pix versus STED, and 0.0441 for CycleGAN versus STED to 0.0293 for PSF-guided AI versus STED. These values indicate that the PSF-guided outputs exhibited fewer abnormal local deviations from the STED reference than the corresponding Pix2Pix and CycleGAN predictions. Together, these qualitative observations support the conclusion that integrating PSF information into the generative framework improves both visual realism and local agreement with the target super-resolved modality.

**Performance Comparison Between PSF-Guided and Non-PSF Models**

To optimize the contribution of the PSF term in the total loss function, we systematically evaluated different PSF-loss weights, αPSF, for both simulated and experimental PSFs and compared them with the non-PSF baseline. The effect of these settings on confocal-to-STED translation was assessed using the STED ground truth, the confocal input, the AI-generated image obtained with PSF guidance at αPSF = 50, and the overconstrained AI-generated image obtained with αPSF = 500, as illustrated in Fig. 4a. Overall, PSF guidance improved image generation relative to the non-PSF model, although the extent of improvement depended on the selected PSF-loss weight and on whether simulated or experimental PSFs were used. For both simulated and experimental PSFs, several intermediate settings outperformed the No-PSF baseline, indicating that the incorporation of imaging-system information benefits translation quality. In the full-reference analysis, the simulated PSF models generally achieved slightly stronger performance than the corresponding experimental PSF models, as demonstrated in Fig. 4b–d.

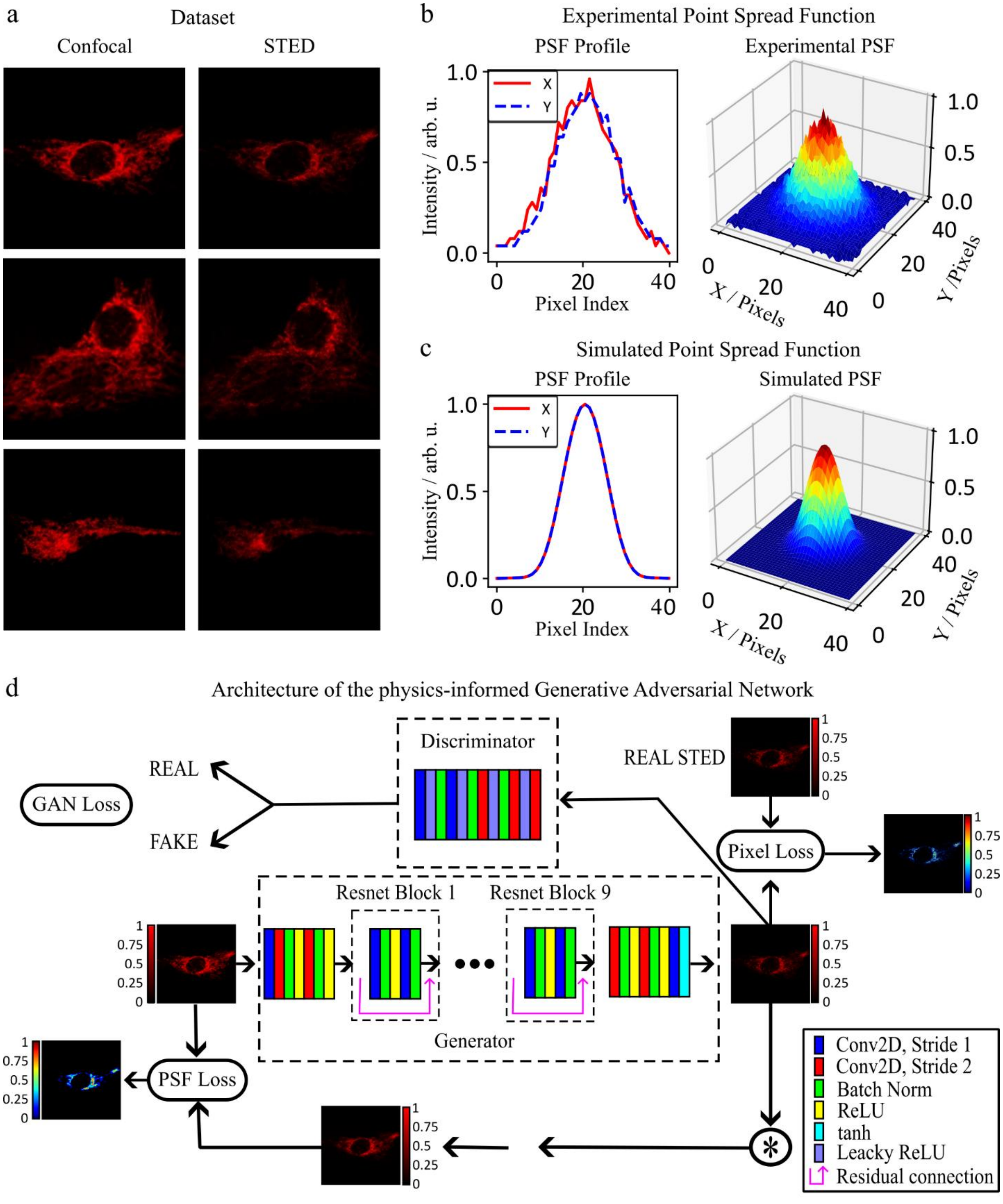

**Fig. 1 a.** Confocal and corresponding STED images of translocase of the outer mitochondrial membrane 20 (TOM20)-labeled cells. **b, c**. Experimental and simulated confocal point spread function (PSF) profiles along the x and y axes and the corresponding 3D PSF representations. **d.** Schematic of the physics-informed generative adversarial network architecture for confocal-to-STED image translation with the PSF-based loss term.

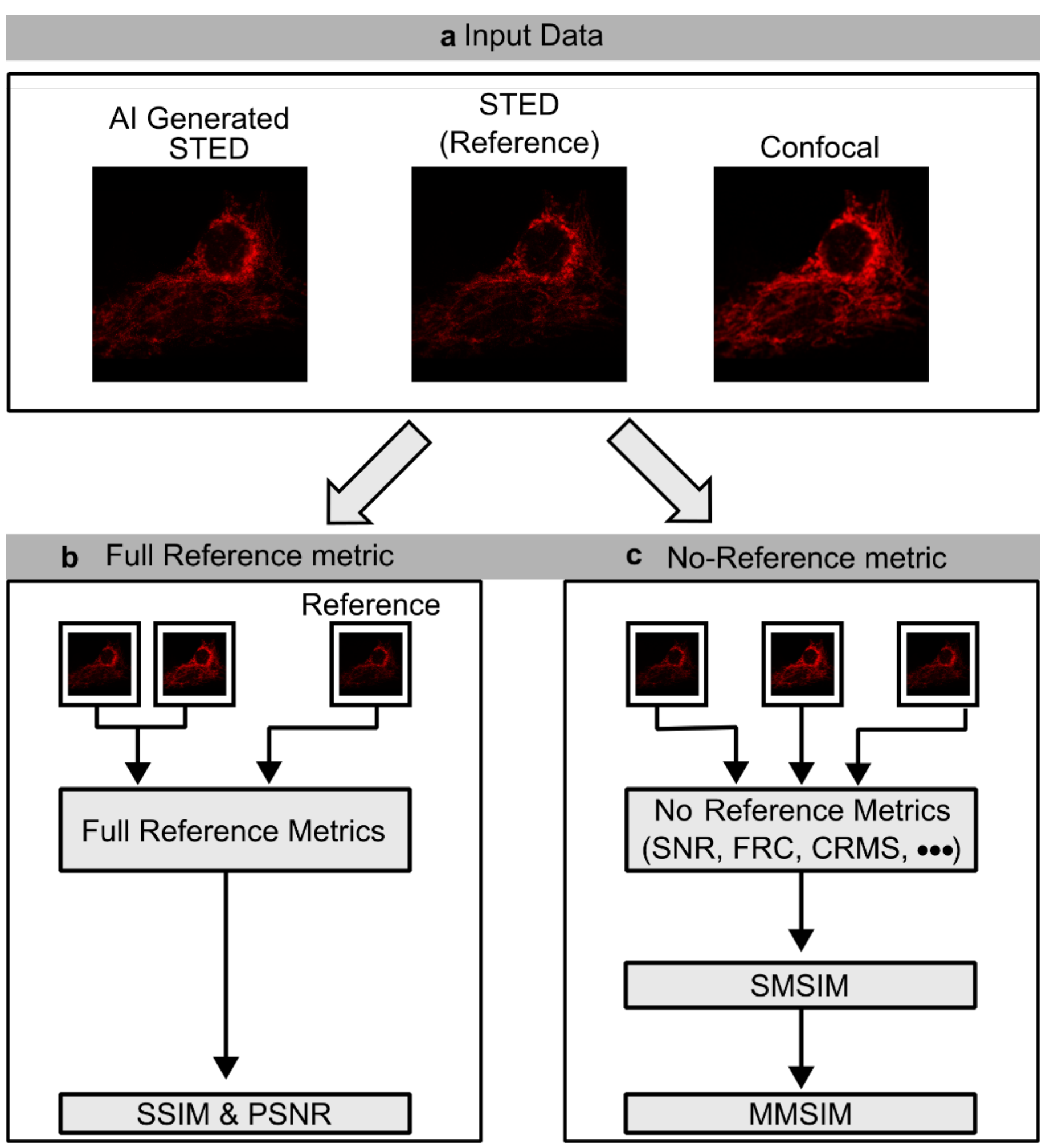

**Fig 2. Schematic of the image-quality analysis pipeline used to compare AI-generated images with STED references. a** Input image data, including the confocal image, the AI-generated image, and the corresponding STED reference image. **b** Full-reference analysis, yielding the structural similarity index measure (SSIM) and peak signal-to-noise ratio (PSNR). **c** Multi-marker similarity analysis, in which no-reference metric vectors are compared with those of the corresponding STED reference to compute single-marker similarity indices, which are then averaged to obtain the multi-marker similarity metric (MMSIM).

The optimal configuration was metric-dependent: the simulated PSF model with $\alpha_{PSF} = 150$ achieved the highest SSIM, followed by the simulated PSF model with $\alpha_{PSF} = 50$, whereas the simulated PSF model with $\alpha_{\mathrm{PSF}} = 50$ achieved the highest PSNR. The highest MMSIM was achieved by the simulated PSF model with $\alpha_{PSF} = 150$, followed closely by the experimental PSF model with $\alpha_{PSF} = 50$. A further important observation is that excessive PSF weighting degraded performance. This deterioration was particularly clear for the simulated $\alpha_{PSF} = 500$ case and was more evident in PSNR and MMSIM than in SSIM. Relative to the non-PSF baseline, PSNR decreased by approximately 12.9% and MMSIM by about 3.7%, whereas SSIM changed by only about 0.07%. Likewise, compared with the STED reference, the confocal input showed a much larger deviation in PSNR and MMSIM than was apparent from SSIM alone. These findings indicate that, in the present setting, PSNR and MMSIM were more sensitive than SSIM to the degradation associated with poor or overconstrained model configurations. A complete numerical summary of all configurations is provided in Table 1.

Table 1. Quantitative mean ± standard deviation of SSIM, PSNR, and MMSIM for all configurations with respect to the STED reference.

| Group | SSIM | PSNR | MMSIM |
|---|---|---|---|
| Confocal | 0.7348 ± 0.0782 | 20.44 ± 1.90 | 0.9316 ± 0.0282 |
| No-PSF | 0.7458 ± 0.0899 | 24.23 ± 1.62 | 0.9745 ± 0.0296 |
| SIM10 | 0.7488 ± 0.0865 | 24.25 ± 1.53 | 0.9771 ± 0.0265 |
| EXP10 | 0.7457 ± 0.0874 | 24.18 ± 1.58 | 0.9752 ± 0.0285 |
| SIM20 | 0.7494 ± 0.0858 | 24.31 ± 1.57 | 0.9760 ± 0.0265 |
| EXP20 | 0.7486 ± 0.0897 | 24.20 ± 1.59 | 0.9768 ± 0.0259 |
| SIM50 | 0.7512 ± 0.0901 | 24.34 ± 1.52 | 0.9786 ± 0.0244 |
| EXP50 | 0.7495 ± 0.0875 | 24.24 ± 1.48 | 0.9803 ± 0.0222 |
| SIM100 | 0.7510 ± 0.0875 | 24.27 ± 1.58 | 0.9762 ± 0.0266 |
| EXP100 | 0.7544 ± 0.0846 | 24.27 ± 1.44 | 0.9786 ± 0.0227 |
| SIM150 | 0.7553 ± 0.0859 | 24.23 ± 1.58 | 0.9804 ± 0.0199 |
| EXP150 | 0.7530 ± 0.0867 | 24.17 ± 1.51 | 0.9791 ± 0.0206 |
| SIM200 | 0.7529 ± 0.0833 | 24.05 ± 1.54 | 0.9788 ± 0.0193 |
| EXP200 | 0.7529 ± 0.0879 | 24.06 ± 1.58 | 0.9794 ± 0.0213 |
| SIM500 | 0.7453 ± 0.0895 | 21.11 ± 3.59 | 0.9382 ± 0.0290 |

## Quantitative Evaluation and Frequency Analysis Based on No-Reference Metrics and Wasserstein Distance

Quantitative evaluation based on no-reference metrics, which characterize image quality without requiring direct comparison to a reference image, further confirmed that AI-generated modalities were substantially closer to the STED reference distribution than the original confocal input, with PSF-guided models providing the strongest agreement in several cases (Fig. 5). These metrics included the mean signal-to-noise ratio (SNRmean), Fourier ring correlation-based resolution (FRCres), root mean square contrast (CRMS), and the summed Fourier ring correlation response (FRCsum), which together capture complementary aspects of image quality, contrast, and frequency content. Because these metrics quantify spatial-frequency content, contrast, and signal-to-noise behavior, they provide a physics-relevant assessment of microscopy image formation beyond pixel-wise similarity. We used the Wasserstein distance to quantify how closely the distribution of each no-reference metric for a given model matched the corresponding STED reference distribution. Unlike comparisons based only on mean values, this approach captures differences across the full distribution and therefore provides a more informative measure of overall agreement with the target modality.

For SNRmean, the confocal input yielded $20.04 \pm 3.87$, whereas STED reached $24.60 \pm 4.28$. Both AI-generated outputs approached or exceeded the STED level, with values of $25.08 \pm 3.95$ for the No-PSF model and $25.01 \pm 4.05$ for the PSF-guided model. Consistently, the Wasserstein distance to the STED distribution was markedly reduced from 4.56 for confocal to 0.524 for No-PSF and 0.516 for PSF-SIM-50, indicating that AI-based translation recovered signal characteristics much closer to the target modality than the original confocal images.

A similar trend was observed for CRMS. Confocal images showed $0.1764 \pm 0.0298$, while STED images exhibited $0.1209 \pm 0.0250$. The AI-generated output shifted strongly toward the STED range, with $0.1174 \pm 0.0209$ for No-PSF and $0.1186 \pm 0.0227$for PSF-SIM-50. The corresponding Wasserstein distances to STED decreased from $5.55 \times 10^{-2}$ for confocal to $4.34 \times 10^{-3}$ for No-PSF and $3.84 \times 10^{-3}$ for PSF-SIM-50, again supporting improved distributional agreement of the PSF-aware model.

The clearest advantage of PSF guidance was observed in the frequency-domain analysis. In microscopy, frequency-domain behavior is particularly important because it reflects the preservation of structurally relevant spatial-frequency information, which is directly linked to the recovery of fine image detail. For FRC-based metrics, the AI-generated outputs were much closer to STED than confocal, and the PSF-aware systems showed the strongest overall agreement. In the boxplot analysis, the mean FRCres and FRCsum values for SIM50 were closer to STED than those of the No-PSF model, indicating improved recovery of STED-like frequency content. This trend was reinforced by the Wasserstein analysis: for FRCres, the distance to STED decreased from 0.2471 for confocal to 0.1247 for No-PSF and further to 0.0231 for SIM50, while the best-performing model, EXP150, reached 0.0106. Paired Wilcoxon signed-rank analysis of image-level FRCres values further showed that the selected PSF-guided models differed significantly from the No-PSF baseline after Benjamini–Hochberg false-discovery-rate correction (Table 2), supporting a reproducible frequency-domain shift associated with PSF guidance. The No-PSF model yielded an FRCres of 0.9855 ± 0.0089, whereas SIM50 reached 0.9987 ± 0.0016 (median difference = 0.0145, $P = 6.93 \times 10^{-10}$, $q = 2.47 \times 10^{-9}$), EXP50 reached 0.9978 ± 0.0027 (median difference = 0.0141, $P = 3.39 \times 10^{-9}$, $q = 7.62 \times 10^{-9}$), EXP150 reached 0.9976 ± 0.0031 (median difference = 0.0133, $P = 3.39 \times 10^{-9}$, $q = 7.62 \times 10^{-9}$), and SIM150 reached 0.9975 ± 0.0031 (median difference = 0.0135, $P = 5.49 \times 10^{-7}$, $q = 6.17 \times 10^{-7}$). These results support a statistically significant frequency-domain shift induced by PSF guidance relative to the non-PSF model.

The heatmap analysis across multiple no-reference metrics showed that PSF-guided systems, particularly SIM50, SIM150, EXP50, and EXP150, consistently yielded smaller Wasserstein distances than the No-PSF baseline, whereas confocal remained the furthest from STED (Fig. 5e; Table 3). The full Wasserstein distance analysis across all no-reference metrics heatmap is provided in Supplementary Fig. S1. In particular, the FRCres analysis showed that PSF-guided models achieved substantially lower Wasserstein distances than both confocal and the No-PSF baseline, with EXP150 giving the closest agreement with STED (Fig. 5f; Table 4). Together, these findings indicate that PSF-aware image generation improves not only overall no-reference image quality but, more importantly, the recovery of STED-like spatial-frequency characteristics. For completeness, boxplot results for all ten no-reference metrics are provided in Supplementary Figs. S2.1–S2.10.

Table 2. Paired Wilcoxon signed-rank analysis of FRCres values relative to the No-PSF baseline.

| | CONF | SIM50 | EXP50 | SIM100 | SIM20 | EXP100 | EXP200 | EXP150 |
|---|---|---|---|---|---|---|---|---|
| **Median Difference** | -0.0155 | 0.0078 | 0.0031 | 0.0055 | 0.0035 | 0.0071 | 0.0039 | 0.0094 |
| **P-values** | 0.016584 | 0.013208 | 0.048123 | 0.049659 | 0.179687 | 0.001388 | 0.011754 | 0.001779 |
| **FDR q** | 0.026535 | 0.026415 | 0.056753 | 0.056753 | 0.179687 | 0.007117 | 0.026415 | 0.007117 |
| **Effect size** | -0.4262 | 0.4402 | 0.3542 | 0.3519 | 0.2427 | 0.5587 | 0.4472 | 0.547 |

Table 3. Wasserstein distances average for the no-reference metrics

| Model | CONF | NO-PSF | SIM50 | SIM150 | EXP50 | EXP150 |
|---|---|---|---|---|---|---|
| **Crms** | 0.1024 | 0.0661 | 0.0352 | 0.0158 | 0.0215 | 0.025 |
| **Eedge** | 0.0403 | 0.0218 | 0.0219 | 0.0166 | 0.0193 | 0.0143 |
| **FRCres** | 0.2471 | 0.1247 | 0.0231 | 0.0271 | 0.0225 | 0.0106 |
| **FRCsum** | 0.2847 | 0.0148 | 0.0092 | 0.0248 | 0.0254 | 0.0264 |
| **HFsum** | 0.1172 | 0.0574 | 0.0416 | 0.0438 | 0.0413 | 0.0378 |
| **LFsum** | 0.0638 | 0.0088 | 0.0086 | 0.0126 | 0.0069 | 0.0093 |
| **MFsum** | 0.147 | 0.0555 | 0.0556 | 0.0717 | 0.0408 | 0.0348 |
| **SC** | 0.0556 | 0.0332 | 0.0319 | 0.0457 | 0.0287 | 0.0335 |
| **SNRmean** | 0.0234 | 0.0171 | 0.0115 | 0.0084 | 0.0136 | 0.0098 |
| **SNRstd** | 0.021 | 0.0202 | 0.0164 | 0.015 | 0.0115 | 0.0183 |

Table 4. Wasserstein distances for FRCres with respect to the STED reference for the top-performing PSF-guided models, together with the confocal input and the No-PSF baseline.

| CONF | NO-PSF | SIM50 | EXP50 | SIM100 | SIM20 | EXP100 | EXP200 | EXP150 |
|---|---|---|---|---|---|---|---|---|
| 0.2471 | 0.1247 | 0.0231 | 0.0225 | 0.0222 | 0.0178 | 0.0172 | 0.0168 | 0.0106 |

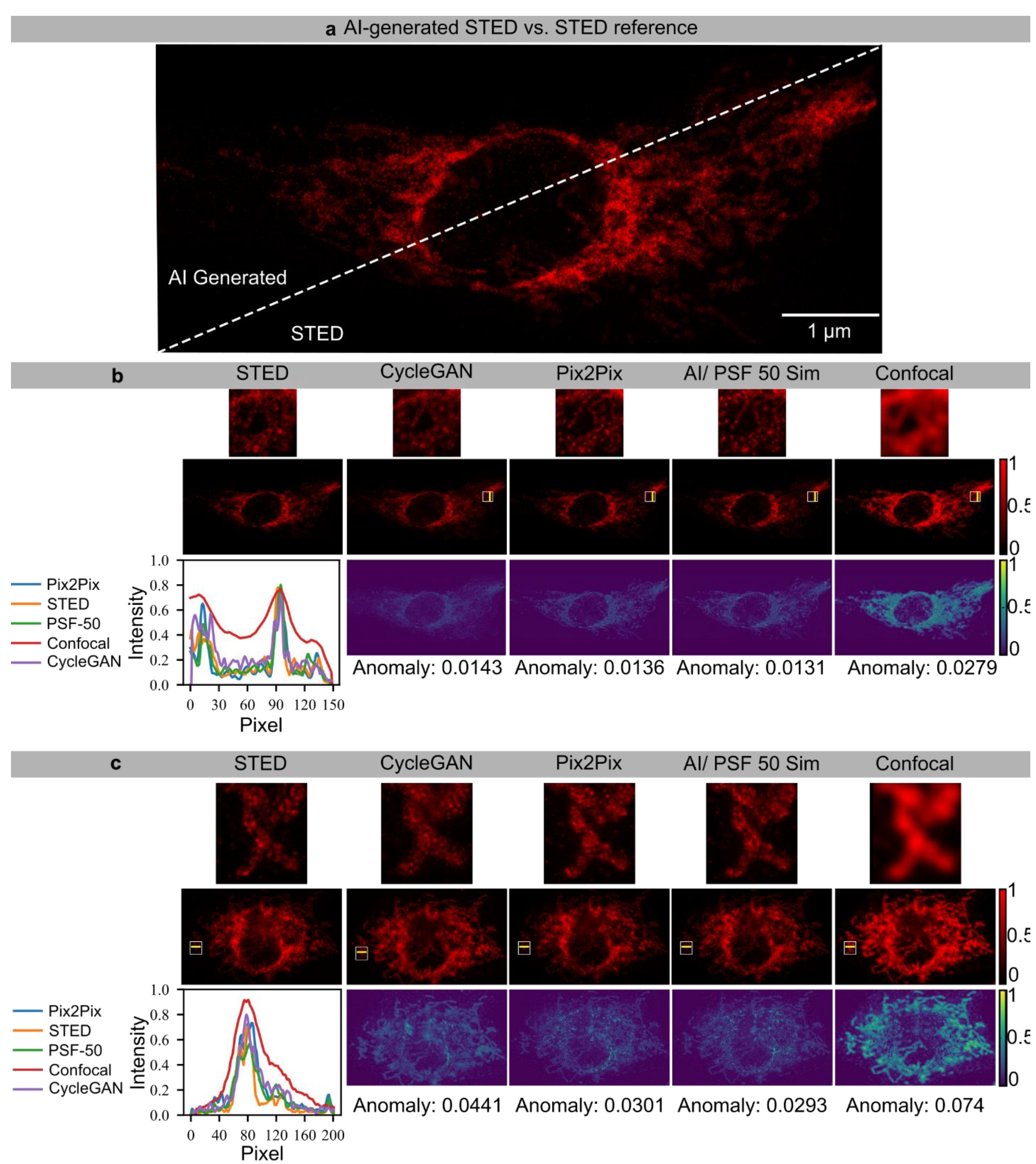

**Fig. 3. Representative comparison of confocal input, AI-generated STED images, and experimentally acquired STED ground truth.**

a, Composite comparison of the AI-generated STED image and the corresponding STED reference. b,c, Two representative fields of view comparing the STED reference, Pix2Pix/No-PSF output, CycleGAN output, PSF-guided output, and confocal input, together with zoomed regions, deviation maps, and line-intensity profiles. White boxes indicate zoomed regions, and yellow lines indicate the positions used for profile extraction. The PSF-guided model shows reduced local deviations and intensity profiles that more closely follow the STED reference, supporting improved local agreement with the target super-resolved modality.

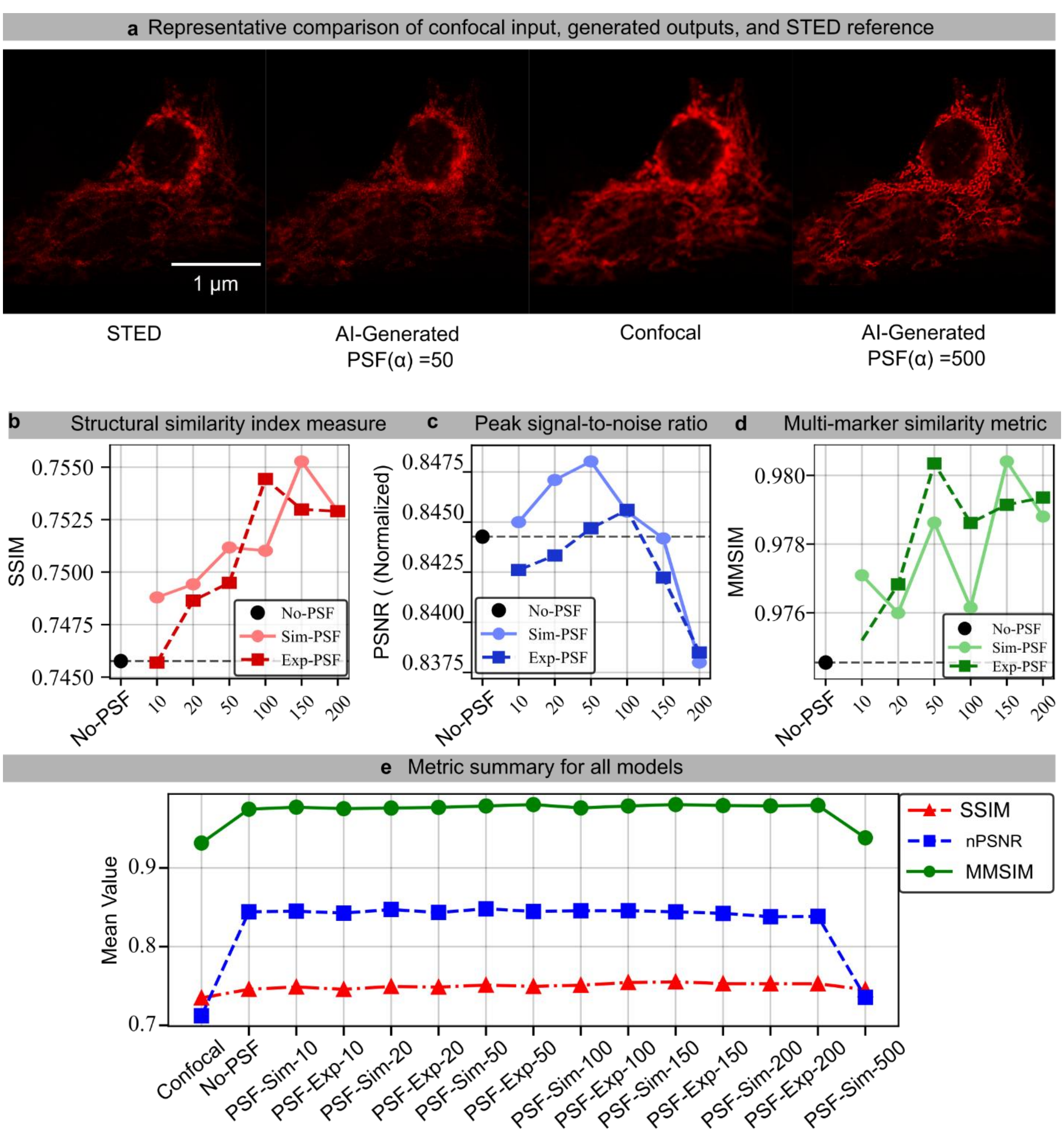

**Fig 4. Quantitative analysis of the effect of point spread function guidance and regularization strength on image-quality metrics. a** Representative image: STED, PSF-guided AI (α = 50), confocal, and PSF-guided AI (α = 500), showing improved reconstruction at α = 50 and degradation at α = 500. **b–d** Structural similarity index measure (SSIM), normalized peak signal-to-noise ratio (PSNR), and multi-marker similarity metric (MMSIM) as a function of point spread function (PSF) regularization strength α for simulated and experimental PSFs, with all models evaluated against the STED ground truth. In all three metrics, PSF (50 and 100) weighting yields better performance than the No-PSF baseline, whereas performance decreases at high α. **e** Mean of SSIM, normalized PSNR, and MMSIM for all models, including the confocal baseline. The results show that the difference between the confocal input and the α = 500 model is proportionally larger for PSNR and MMSIM than for SSIM.

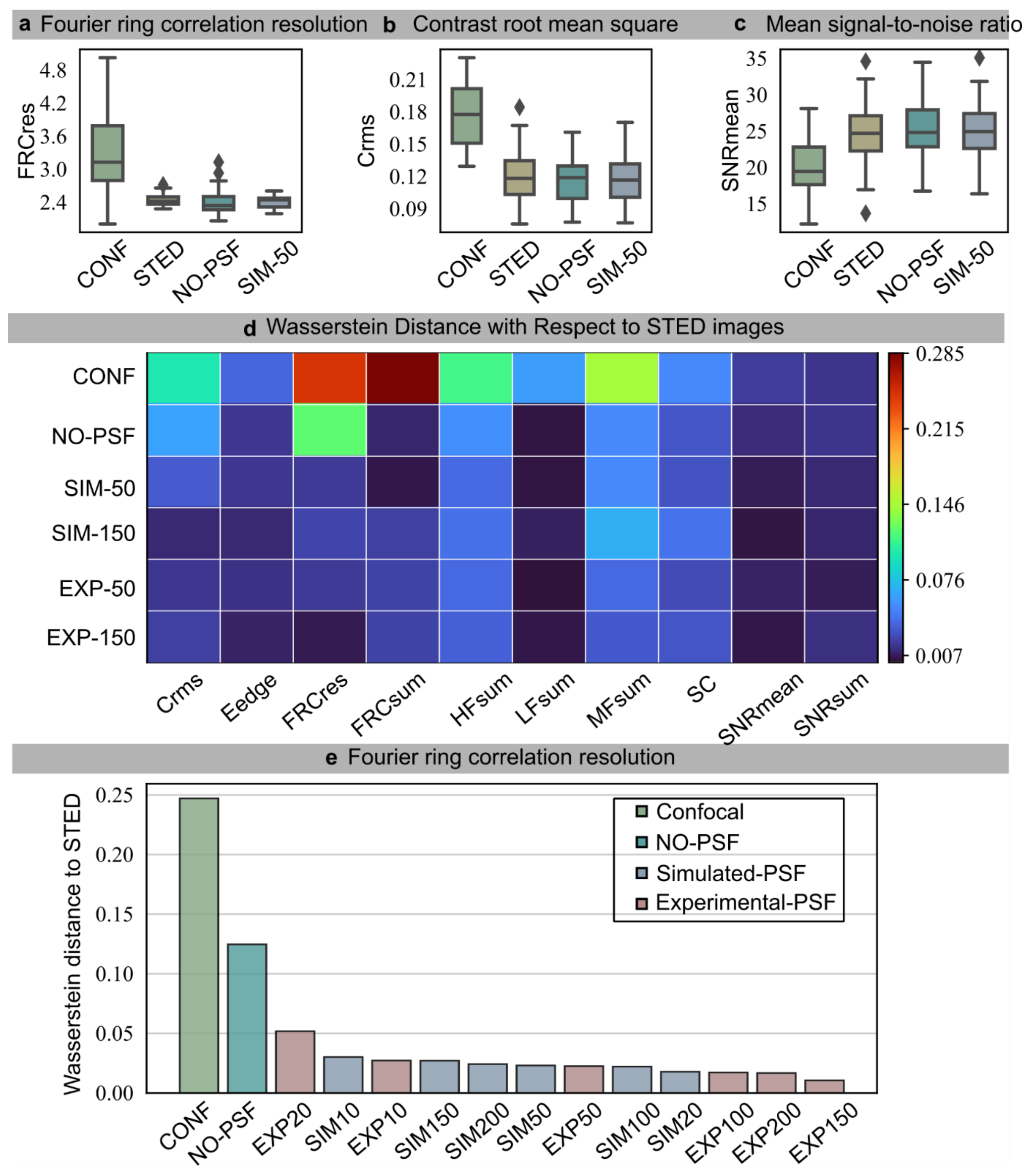

**Fig. 5. Quantitative analysis of no-reference image-quality metrics and frequency-domain agreement with the STED reference. a–c.** Boxplots of Fourier ring correlation-based resolution (FRCres), contrast root mean square (CRMS), and mean signal-to-noise ratio (SNRmean) for confocal, STED, No-PSF, and PSF-guided images with simulated PSF guidance at α = 50. The distributions for the PSF-guided model are generally closer to STED than those of the No-PSF model, with the clearest improvement in FRCres. **d.** Heatmap of normalized Wasserstein distances for selected no-reference metrics relative to STED, where smaller values indicate closer agreement with the STED distribution. **e.** Wasserstein ranking for FRCres, showing that PSF-guided models with simulated and experimental PSFs achieve closer frequency-domain agreement with the STED reference than the confocal input and the No-PSF model, consistent with improved recovery of fine spatial detail.

## Discussion

The present results show that incorporating point spread function (PSF) information into the loss function improves confocal-to-STED microscopy image translation beyond a purely data-driven baseline. The advantage of PSF awareness became particularly evident in frequency-domain and distribution-based analyses. This is important in microscopy because the fidelity of generated images should not be judged only by global structural similarity, but also by whether the recovered image statistics and spatial-frequency content remain consistent with the target modality. Thus, the no-reference metrics used here provide a physics-relevant complement to conventional similarity scores. Because the paired images were acquired across different experimental days, the evaluation also partially probes robustness to day-to-day experimental variability.

A key observation is that PSF guidance improves the physical consistency of the translation process. By constraining the generated STED microscopy-like output through the confocal image-formation model, the network is encouraged to produce images that are not only visually plausible but also compatible with the measured lower-resolution input. This likely explains the reduced local deviations, lower anomaly scores, and improved agreement with the STED images in several no-reference and frequency-sensitive metrics.

The frequency-domain results are particularly relevant. In fluorescence microscopy, preservation of spatial-frequency information is closely linked to the recovery of fine structural detail. The FRC-based analyses and the associated Wasserstein distances showed that PSF-aware systems were more consistent with the STED reference than both the confocal input and the non-PSF baseline. This indicates that the proposed framework does not merely sharpen images, but more effectively recovers the statistical and frequency-domain characteristics of the target super-resolved modality. The improvement in FRC-based metrics indicates that the PSF constraint does not simply regularize the GAN output, but enforces consistency with the confocal forward model, thereby suppressing frequency components that are incompatible with the measured input while preserving STED-like high-frequency structure.

The best-performing model depended on the evaluation metric. The simulated PSF model with $\alpha_{PSF} = 150$ achieved the highest SSIM and MMSIM, whereas the highest PSNR was obtained for the simulated PSF model with $\alpha_{PSF} = 50$. For the FRCres Wasserstein analysis, the best agreement with STED was obtained for the experimental PSF model with $\alpha_{\mathrm{PSF}} = 150$. This emphasizes that no single metric fully captures microscopy image quality and supports the combined use of reference-based, no-reference, and distribution-based evaluation.

The over-constrained $\alpha_{\mathrm{PSF}} = 500$ setting showed that PSNR and MMSIM were more sensitive than SSIM to quality degradation, highlighting the value of microscopy-oriented no-reference metrics. This is especially relevant because in many practical imaging settings, paired ground-truth data are limited or unavailable. Future work could extend the PSF-aware framework to unpaired GAN systems, where physics-informed losses may provide an alternative source of supervision and help stabilize modality transfer in the absence of paired training data. The approach could also be adapted to multichannel fluorescence imaging and extended to multichannel image-fusion tasks, where physically grounded constraints may help preserve biologically relevant structural relationships across channels. In addition, applying the framework to other microscopy modalities, three-dimensional imaging, and cross-instrument translation may further clarify its generalizability and practical value. Beyond image generation, the framework may also support experimental image-quality assessment, as demonstrated in our previous work (37). More broadly, integrating optical priors into generative models may provide a versatile strategy for improving both the realism and physical plausibility of AI-based image reconstruction in fluorescence microscopy.

## Materials and Methods

### Sample Preparation

Human M2 macrophages were generated from leukocyte concentrates from healthy blood donors, the general protocol is described in Günther et al (42). M2 macrophages were then immunolabeled for Translocase of the Outer Mitochondrial Membrane 20 (TOM20), a receptor of the mitochondrial outer membrane import machinery that is widely used as a mitochondrial marker (38). After fixation, M2 macrophages were fluorescently labeled using a rabbit polyclonal anti-TOM20 primary antibody (11802-1-AP, Proteintech; 1:100), followed by a goat anti-rabbit STAR RED secondary antibody (STRED-1002, Abberior; 1:200), according to standard immunofluorescence protocol. Fluorescence labeling was performed using Abberior STAR RED, a far-red dye compatible with confocal and STED microscopy and commonly operated with 640 nm excitation (39). The dataset comprised paired confocal and STED images acquired across different experimental days from TOM20-labeled mitochondria and was used for confocal-to-STED modality transfer experiments; representative sample images are shown in Fig. 1a.

**Microscope Setup and Image Acquisition**

Confocal input images were acquired using a laser scanning fluorescence microscope operated in confocal mode with an oil-immersion objective of numerical aperture 1.4. The refractive indices of the immersion and embedding media were both 1.5. Image sampling in the lateral directions was 20 nm, and a back-projected pinhole of 250 nm was used for confocal acquisition. Corresponding confocal and STED images were recorded from the same fields of view to generate paired data for supervised modality transfer. These paired measurements enabled direct comparison between the experimentally acquired STED reference images and the AI-generated outputs (40).

**Simulated and Experimental PSFs**

To incorporate imaging-system knowledge into the proposed framework, both simulated and experimentally measured point spread functions (PSFs) were used. The simulated PSF was generated with Huygens Professional using the optical parameters of the confocal imaging configuration and served as an idealized estimate of the system response. In addition, an experimental PSF was measured from GATTA-Beads acquired under matching imaging conditions. GATTA-Beads are fluorescent calibration beads with a reported diameter of 23 nm, which makes them suitable for PSF characterization in super-resolution microscopy. The measured PSF, therefore, captures practical system-specific effects that are not fully represented by simulation alone. Both PSF types were integrated into the proposed framework to evaluate the effect of physics-guided constraints on confocal-to-STED modality transfer. Representative examples of the simulated and experimental PSFs are shown in Fig. 1b and Fig. 1c.

**PSF-Aware GAN Framework**

To translate confocal images into STED-like representations, we employed a supervised conditional generative adversarial network (cGAN) augmented with a physics-guided point spread function (PSF) consistency term. The weighting coefficient of the PSF-consistency term was optimized to determine the most effective balance between adversarial learning, supervised reconstruction, and imaging-physics guidance.

The generator G was implemented using a ResNet architecture with nine residual blocks, while the discriminator D followed a PatchGAN design that evaluates local image patches and promotes realistic high-frequency structure. As in conditional image translation, the adversarial objective is written as:

$$\mathcal{L}_{\text{cGAN}}(G, D) = \mathbb{E}_{x,y}\left[Log\ D(x, y)\right] + \mathbb{E}_{x}\left[Log\left(1 - D\left(x, G(x)\right)\right)\right] \quad (1)$$

where $x$ denotes the input confocal image, $y$ the corresponding target STED image, $G(x)$ the generated STED-like output, and $D(x,\cdot)$ the discriminator conditioned on the confocal input. Here, $\mathbb{E}_{(x\cdot y)}$ and $\mathbb{E}_{x}$ denote expectations, i.e., averaging over the paired training samples and the input-image distribution, respectively.

To enforce similarity between the generated image and the target STED image, we used a pixel-wise reconstruction term (11):

$$\mathcal{L}_{\text{pix}}(G) = \mathbb{E}_{x,y}\left[\| y - G(x) \|_1\right] \quad (2)$$

To further constrain the solution by the imaging process, we introduced a PSF consistency loss. Specifically, the generated STED-like image was convolved with the confocal PSF and compared with the measured confocal input image,

$$\mathcal{L}_{\text{PSF}}(G) = \mathbb{E}_{x}\left[\| h_{\text{conf}} * G(x) - x \|_1\right] \quad (3)$$

where $h_{\text{conf}}$ is the confocal PSF and $*$ denotes convolution. For computational efficiency, this operation was implemented in the Fourier domain according to the convolution theorem, i.e., by multiplying the Fourier transform of $G(x)$ with the Fourier transform of $h_{\text{conf}}$, followed by an inverse transformation to the spatial domain.

The final objective is therefore given by:

$$G^* = \arg\min_G \max_D [\mathcal{L}_{\mathrm{cGAN}}(G, D) + \mathcal{L}_{\mathrm{pix}}(G) + \alpha_{\mathrm{PSF}}\, \mathcal{L}_{\mathrm{PSF}}(G)] \qquad (4)$$

where $\alpha_{\mathrm{PSF}}$ denotes the weighting coefficient for the PSF-consistency loss. In this study, $\alpha_{\mathrm{PSF}}$ was systematically optimized to determine the most effective balance between supervised reconstruction and physics-guided consistency. The network architecture is shown in Fig. 1d. The Pix2Pix model was used as the supervised non-PSF baseline, using the same paired confocal–STED training data and the same adversarial and pixel-wise reconstruction losses, but without the PSF-consistency term. CycleGAN was included as an additional image-translation baseline to compare against an unpaired translation framework. Both baseline models were evaluated using the same visual comparison, deviation-map, line-profile, and anomaly-score analysis used for the PSF-guided model.

**Training and Validation Protocol**

The dataset comprised 42 paired confocal and STED images of TOM20-labeled mitochondria acquired on different experimental days. Model performance was evaluated using five-fold cross-validation, in which the dataset was partitioned into five subsets. In each fold, four subsets, corresponding to approximately 33–34 image pairs, were used for training, and the remaining subset, corresponding to 8–9 image pairs, was used for testing. To improve reproducibility across experiments, the training environment was reset before each run by clearing GPU memory, synchronizing CUDA operations, resetting random seeds for Python, NumPy, and PyTorch, and performing garbage collection.

**Evaluation Metrics**

To comprehensively assess the quality of the generated STED-like images, we employed both full-reference and no-reference image quality metrics, as summarized in Fig. 2. The full-reference evaluation was based on the structural similarity index measure (SSIM) and the peak signal-to-noise ratio (PSNR), both computed between the AI-generated STED images and the experimentally acquired STED ground truth. These metrics quantify structural agreement and pixel-level fidelity with respect to the reference target.

To complement this analysis, we further applied the multi-marker similarity metric (MMSIM) as a full-reference evaluation method. MMSIM is calculated from 10 underlying no-reference image-quality metrics, but in the present implementation, it serves as a reference-based metric because the resulting marker values are compared with those of the ground-truth (GT) reference. Specifically, ten no-reference metrics were computed: Fourier ring correlation resolution (FRCres), Fourier ring correlation sum (FRCsum), signal-to-noise ratio standard deviation (SNRstd), signal-to-noise ratio mean (SNRmean), high-frequency sum (HFsum), low-frequency sum (LFsum), mid-frequency sum (MFsum), structural complexity (SC), edge count measure (Eedge), and contrast root mean square (CRMS). These descriptors capture complementary aspects of image quality, including frequency-dependent information, signal-to-noise characteristics, edge content, and contrast-related properties. The individual no-reference measures were first combined into a summarized score denoted as single-marker similarity metrics (SMSIM). Subsequently, this score was incorporated into the previously introduced MMSIM framework, which provides an aggregated image-quality estimate based on multiple microscopy-relevant descriptors and is particularly useful in situations where no ground-truth reference image is available (41).

By combining full-reference and no-reference evaluation, the analysis enables assessment of both agreements with experimentally acquired STED images and intrinsic image-quality characteristics that are not fully captured by pixel-wise comparison alone. This multi-metric strategy was used throughout the study to compare confocal input images, generated outputs, and experimentally measured STED references within a unified evaluation framework.

**Statistical analysis**

To assess whether PSF-guided models differed significantly from the No-PSF baseline, image-level FRCres values were compared using paired Wilcoxon signed-rank tests. The tests were performed on paired measurements from the same images, and median differences were calculated as model minus No-PSF. Because multiple model comparisons were performed, P values were adjusted using the Benjamini–Hochberg false-discovery-rate procedure. Corrected values are reported as q values, with $q < 0.05$ considered statistically significant. Effect sizes were reported using the rank-biserial correlation (Table 2) (43, 44).

**Acknowledgements**

This work is supported by the BMFTR, funding program Photonics Research Germany (13N15715 (LPI-BT4-FSU), 13N15719 (LPI-BT5-FSU), 13N15710 (LPI-BT3-FSU), LPI-BT4-Leibniz-IPHT (13N15713), LPI-BT5-Leibniz-IPHT (13N15717)) and is integrated into the Leibniz Center for Photonics in Infection Research (LPI). The LPI initiated by Leibniz-IPHT, Leibniz-HKI, Friedrich Schiller University Jena and Jena University Hospital is part of the BMBF national roadmap for research infrastructures. Deutsche Forschungsgemeinschaft (DFG, German Research Foundation; Germany´s Excellence Strategy – EXC 2051 – Project-ID 390713860; project number 316213987 – SFB 1278; GRK M-M-M: GRK 2723/1 – 2023 – ID 44711651; RTG 3014 "PhInt - Photo-Polarizable Interfaces and Membranes" Project number 521747072; Instrument funding modular STED INST 1757/25-1 FUGG), the State of Thuringia (TMWWDG), the Leibniz Association (Leibniz Collaborative Excellence Programme, project AMPel – project number K548/2023), the ESF Plus and the Free State of Thuringia (TAB; Advanced Flu-Spec / 2020 FGZ: FGI 0031), SFB1127/3ChemBioSysproject 239748522. A part of the project on which these results are based was funded by the Free State of Thuringia under the number 2018 IZN 0002 (Thimedop) and co-financed by funds from the European Union within the framework of the European Regional Development Fund (EFRE). The work is co-funded by the European Union (ERC, STAIN-IT, 101088997). Views and opinions expressed are however those of the author(s) only and do not necessarily reflect those of the European Union or the European Research Council. Neither the European Union nor the granting authority can be held responsible for them.

**Author contributions**

M.S. developed the artificial intelligence models and implemented the corresponding code. M.S. and E.C. developed the data-analysis code, and M.S. and T.B. analyzed the data. M.S. and T.B. wrote the manuscript. E.C. developed the multi-marker metrics method and implemented the corresponding code. P.M.J. and O.W. prepared the samples. F.P.L. and C.E. provided the confocal and STED images.

**Data Availability**

The data supporting the findings of this study are available at: https://cloud.uni-jena.de/s/MELZtrtjZ9q3wEK, under datasets/my_data/data_PIDL.

**Code Availability**

The source code used in this study is available at: https://cloud.uni-jena.de/s/MELZtrtjZ9q3wEK.Conflict of interest

The authors declare no competing interests.

**Supplementary information**